\documentclass[a4paper, 10pt, conference]{ieeeconf}     

\IEEEoverridecommandlockouts                             

\usepackage[final]{graphics} 
\usepackage{epsfig} 
\usepackage{mathptmx} 
\usepackage{times} 
\usepackage{amsmath} 
\usepackage{amssymb} 
\usepackage{gensymb}
\usepackage[dvipsnames]{xcolor}
\usepackage{multicol}
\usepackage{ctable}
\usepackage{url}

\usepackage{subcaption}

\title{\LARGE \bf
Road scenes analysis in adverse weather conditions by polarization-encoded images and adapted deep learning }

\author{Rachel Blin$^{1}$, Samia Ainouz$^{1}$, St\'ephane Canu$^{1}$ and Fabrice Meriaudeau$^{2}$
\thanks{$^{1}$The authors are with Normandie Univ, INSA Rouen, UNIROUEN, UNIHAVRE, LITIS,
76000 Rouen, France
        {\tt\small \{rachel.blin, samia.ainouz, stephane.canu\}@insa-rouen.fr}}%
\thanks{$^{2}$Fabrice Merieudeau is with University of Burgundy, UBFC, ImViA, 71200 Le Creusot, France
        {\tt\small fabrice.meriaudeau@u-bourgogne.fr}}%
}

\begin{document}

%

\maketitle
\thispagestyle{empty}
\pagestyle{empty}

\begin{abstract}

Object detection in road scenes is necessary to develop both autonomous vehicles and driving assistance systems. Even if deep neural networks for recognition task have shown great performances using conventional images, they fail to detect objects in road scenes in complex acquisition situations. In contrast, polarization images, characterizing the light wave, can robustly describe important physical properties of the object even under poor illumination or strong reflections. This paper shows how  non-conventional polarimetric imaging modality overcomes the classical methods for object detection especially in adverse weather conditions. The efficiency of the proposed method is mostly due to the high power of the polarimetry to discriminate any object by its reflective properties and on the use of deep neural networks for object detection. Our goal by this work, is to prove that polarimetry brings a real added value compared with RGB images for object detection. Experimental results on our own dataset composed of road scene images taken during adverse weather conditions show that polarimetry together with deep learning can improve the state-of-the-art by about 20\% to 50\% on different detection tasks. 
\end{abstract}

\section{INTRODUCTION}

Road scene understanding is a vital task nowadays because of the development of driving assistance systems. To get a secure navigation and avoid correctly obstacles in road scenes, it is important to get a robust detection. As one knows, it is primordial to ensure a guarantee of safety when talking about autonomous cars because the slightest dysfunction can lead to serious consequences implying human life. Currently, in ideal cases (i.e. good weather and good visibility), road scenes obstacles are well detected. An example of such systems are Mobileye \cite{yoffie2014mobileye} or Waymo \footnote{More information can be found at: \url{https://waymo.com/}}
that achieve a high detection accuracy in such ideal cases. However, when there's variation of illumination or adverse weather conditions leading to unstable appearances in road scenes, most of the methods of the literature implying conventional vision sensors fail to efficiently detect road objects such as vehicles or pedestrians. 
Several methods using conventional sensors have been developed to perform a better detection in road scenes. A contrast restoration approach have been introduced by Hauti\`ere and al. \cite{hautiere2014enhanced}, using a classical RGB camera in order to improve free-space detection in adverse weather conditions. In the same context, Babari and al. \cite{babari2012visibility} proposed an estimation of fog density to enhance object detection and visibility distance using conventional roadside cameras.

Even if all those methods contributed to improve object detection in road scenes, they also demonstrated the limits of using classical imaging sensors. Non-conventional sensors which bring additional features have been introduced in the autonomous driving field to overcome the detection problems occurring with conventional systems. For instance, infrared imaging enabled Miron and al. \cite{miron2015evaluation} to propose an enhanced pedestrian classification system. Infrared imaging was also used by Bertozzi and al. \cite{bertozzi2006low} for their proposed tetra-vision system for pedestrian detection. 

Meanwhile polarization imaging was gaining popularity in other areas including 3D reconstruction \cite{morel2006active}, bio-medical imaging analysis \cite{novikova2018mueller} and military applications \cite{rankin2010passive}. To our knowledge, there are few works that attempt to use polarization for road scene object detection \cite{fan2018polarization}, \cite{kamann2018automotive}.

The principle of the polarization-encoded imaging is that it characterizes the reflected light wave from any object of the scene. Polarization is able to describe important physical properties of the object including its surface geometric structure, the nature of its material as well as its roughness \cite{wolff1995polarization}, \cite{blanchon2019outdoor} even under poor illumination or strong reflections. The polarization state of the reflected light is highly related to the physical properties of an object such as its intensity, its shape and its reflection properties. It is important to know that polarization was applied in several fields  However, this work, is the first one detecting road scene objects in adverse weather conditions.

Deep neural networks have demonstrated their efficiency regarding the object detection in an image. Those networks not only can detect an object but also achieve to make it really fast by processing several images per second. Girshick and al. \cite{girshick2014rich} proposed R-CNN, a region-based convolutional neural network that was able to detect objects in an image. It was the first network able to detect the region containing an object while being able to classify the object. This network then evolved to Fast R-CNN \cite{girshick2015fast} and to Faster-R-CNN \cite{ren2015faster} with both an improvement in accuracy and processing time. As a matter of fact, its processing time has a frame rate of 5fps and it achieved state-of-the-art object detection accuracy on PASCAL VOC 2007 \cite{everingham2010pascal}. This processing time was further improved by Redmond and al. \cite{Redmon_2016_CVPR}  with YOLO. But even though it could detect objects in images at a frame rate of 45 fps, its accuracy couldn't achieve the one of Faster-RCNN. Since YOLO, Liu and al. \cite{liu2016ssd} proposed SSD, a single shot multibox detector that outperformed Faster R-CNN object detection accuracy on PASCAL VOC 2007 but with a higher frame rate of 16fps. More recently, Lin and al. \cite{Lin2017FocalLF} outperformed SSD detection accuracy on PASCAL VOC 2007 with RetinaNet. RetinaNet's frame rate is 14fps which is slightly lower than SSD's but the network achieves a higher performance in small objects detection. By the time, Redmond and al. improved YOLO since its first version to YOLOv2 \cite{redmon2017yolo9000} that outperformed SSD's object detection accuracy on PASCAL VOC 2007 with a frame rate of 40fps, and recently released YOLOv3 \cite{redmon2018yolov3}. The accuracy of all those networks as well as their processing time make them a major asset for object detection in road scenes as well as their deployment in the field of autonomous vehicles. Most of those networks have shown their efficiency for object detection in road scenes, using the KITTI dataset \cite{geiger2012we}, \cite{janai2017computer}.

We aim in this paper to combine the power of polarization to discriminate objects and deep neural networks to detect road scene content even in poor illumination conditions. The idea of using deep learning is motivated by our previous work that show how polarimetry may contribute efficiently to road scene analysis \cite{fan2018polarization} by using only classical machine learning methods (DPM, HOG). Thanks to their high accuracy for object detection, RetinaNet and SSD are chosen to reach our goals. Moreover, due to the lack of polarization road scene dataset, we constituted our own dataset in different weather conditions in Rouen City, France. Experiments show the positive impact of the combination of polarimetry and deep neural networks.

\section{POLARIZATION FORMALISM AND MOTIVATIONS}

Before giving further details regarding the work carried out in this paper, it is important to specify a few polarimetry notions.

\subsection{POLARIZATION FORMALISM}

Polarization is a property  of light waves that can oscillate with more than one orientation. It represents the direction in which the wave is travelling. There exist three states of polarization of a light wave: totally polarized (i.e. the direction of the wave is well determined (elliptic, linear or circular)), unpolarized where the wave has random direction and partially polarized where the wave is a combination of two parts: a totally polarized part and an unpolarized part \cite{bass1995handbook}. Polarimetric imaging is the representation of the polarization state of the light wave reflected from each pixel of the scene. It is mainly used to dissociate metallic object from dielectric surface \cite{born2013principles}. When an unpolarized light wave is being reflected, it becomes partially linearly polarized depending on the surface normal and on the refractive index of the material. That reflected light wave can be described by a measurable set of parameters, the linear Stokes vector S = [$S_0$, $S_1$, $S_2$] where $S_0>0$ is the object total intensity whereas $S_1$ and $S_2$ describe roughly the amount of a linearly polarized light. 

From the Stokes parameters, other physical properties may result such as the angle of polarization (AOP) and the degree of polarization (DOP) \cite{ainouz2013adaptive}. Polarization images are obtained with a polarizer oriented at a specific orientation angle  placed between the scene and the sensor. In order to get the three Stokes parameters, at least three acquisitions with three different polarizer orientations are required. The polarimetric camera used for this work is of the range of Polarcam 4D Technology. It enables to get simultaneously four images, each being obtained with linear polarizer placed at 4 different angles $(\alpha_i)_{i=1:4}$ (0\degree, 45\degree, 90\degree and 135\degree). For each angle $(\alpha_i)$, the camera measures an intensity $I(\alpha_i)$ for the scene. The relationship between the Stokes parameters and the intensities, for each $(\alpha_i)_{i=1:4}$, measured by the camera is given by:

$$
    I(\alpha_i) = \frac{1}{2}[1, \cos(2\alpha_i), \sin(2\alpha_i)].[S_0, S_1, S_2]^T.
$$

The AOP and DOP can be determined from the obtained Stokes vector by:

\vspace*{-.75cm}

\begin{multicols}{2}
$$ 
\qquad     AOP = \frac{1}{2}\tan^{-1}\left(\frac{S_2}{S_1}\right),
$$ 
\vfill
\columnbreak
$$ 
    DOP = \frac{\sqrt{S_1^2+S_2^2}}{S_0}.
$$ 
\end{multicols}
The $DOP \in [0,1]$ refers to the quantity of  polarized light in a wave. It is up to one for the totally polarized light and to zero for unpolarized light. The $AOP \in \left[\frac{-\pi}{2};\frac{\pi}{2}\right]$ is the orientation of the polarized part of the wave with regards to the incident plan

\subsection{MOTIVATIONS}
In our previous work \cite{fan2018polarization}, we demonstrated that by a convenient fusion scheme, polarization features with RGB ones achieve a higher performance for car detection purpose. In this work, a feature selection is performed to select the most informative one among five relevant polarization features. It was found that AOP is the most relevant informative feature. An AOP-based DPM detector (polar-based model) and a color-based DPM detector (color-based model) are then trained independently. Different score maps are produced by the two models. A fusion rule that takes the polar-based model as a confirmation (AND-fusion) of the color-based one to produce the final detection bounding boxes was performed. Experiments proved that taking the complementary information provided by the polarization feature reduces largely the false alarm rate (false bounding boxes), and improve the detection accuracy.

Going from these first encouraging results, this paper shows the effects of the most recent methods, based on deep learning frameworks, for polarization-based object detection purpose, including cars and pedestrians. This work aim to demonstrate that polarization is beyond just a rich panel of color information. The physical information provided by this modality is learned by the used deep architectures and outperform the classical detection methods. 

\section{METHODOLOGY}

Achieving a strong and reliable learning requires both numerous and truthful data. As available polarimetric data are scarce, it was decided to acquire some new polarization images in real scenarios to complement existing RGB databases as explained below. 

This work shows the performances of both RetinaNet and SSD pre-trained on the MS COCO RGB dataset \cite{lin2014} and fine tuned on different polarization channels combinations; Intensities related to polarization angles ($I_0$, $I_{45}$, $I_{135}$), Stokes vector ($S_{0}$, $S_{1}$, $S_{2}$) and ($S_0$, DOP, AOP). The weights of RetinaNet-50 (which refers to RetinaNet using ResNet50 \cite{He_2016_CVPR} as a backbone) and SSD300 (which refers to SSD taking a 300x300 input) using VGG16 \cite{simonyan2014very} as a backbone are kept fix for these new input images to evaluate the results at a first stage. Fine tuning has been then achieved on a 2730 polarization images database \footnote{The dataset and the weights of RetinaNet-50 fine tuned on it can be found at: \url{http://pagesperso.litislab.fr/rblin/databases/}} to prove how Polarization-based detection overcomes the RGB-based one.

\subsection{DATA ACQUISITION}
In order to diversify the images, the acquisition were made while driving to get the more realistic possible images scenario. Following the methodology of the Berkeley Deep Drive database \cite{yu2018bdd100k}, a polarimetric camera was mounted  behind the windshield at the height of the driver eyes in order to make real time acquisitions representing what the driver actually sees in his car. By doing so, we were able to get a large diversity of road scenes which enabled us to avoid over-fitting when training the network. 

It is important to note that a rig made of a RGB camera placed next to the polarimetric camera was also used to acquire images that are used for the testing. By doing so, we are able to test our network on the exact same scenes on two different modalities (RGB, polarimetry).

We would like to stress on the point that the training data were acquired under sunny weather conditions during winter whereas the testing data were taken under foggy conditions in autumn. 

\subsection{DATA SORTING AND LABELLING}
Once enough data were collected, it was important to sort those data in order to maximize the diversity of the database. Our polarimetric camera has a frequency of 15 frames per second so, knowing that acquisitions took place mostly in cities where the speed limit is relatively low and the car had to stop briefly due to road signals, it was decided to keep 1 out of 25 frames. By doing so, a trade-off was achieved, the resemblance between two successive images of the database was minimized while maximizing the number of images in the database. 

\begin{figure}
\centering
\includegraphics[width=0.47\textwidth]{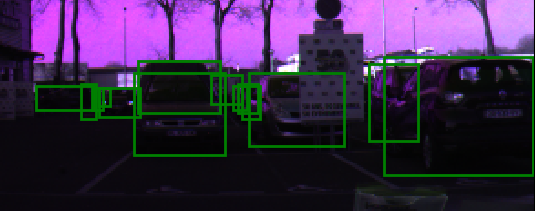}
\caption{Labelling of the objects in a parking}
\label{fig:1}
\end{figure}

Afterwards each image was labelled by means of bounding boxes. 4 categories of object (car, person, bike, motorbike), which are the most common objects present in road scenes were used. Every object was labelled in our data-set, including semi occluded objects such as cars behind obstacles, mostly occluded objects such as parts of windshields corresponding to cars parked behind many others in car parks and small objects such as cars far away. Figure~\ref{fig:1} illustrates the level of precision of this labelling. 

\begin{figure}
\centering
\includegraphics[width=0.47\textwidth]{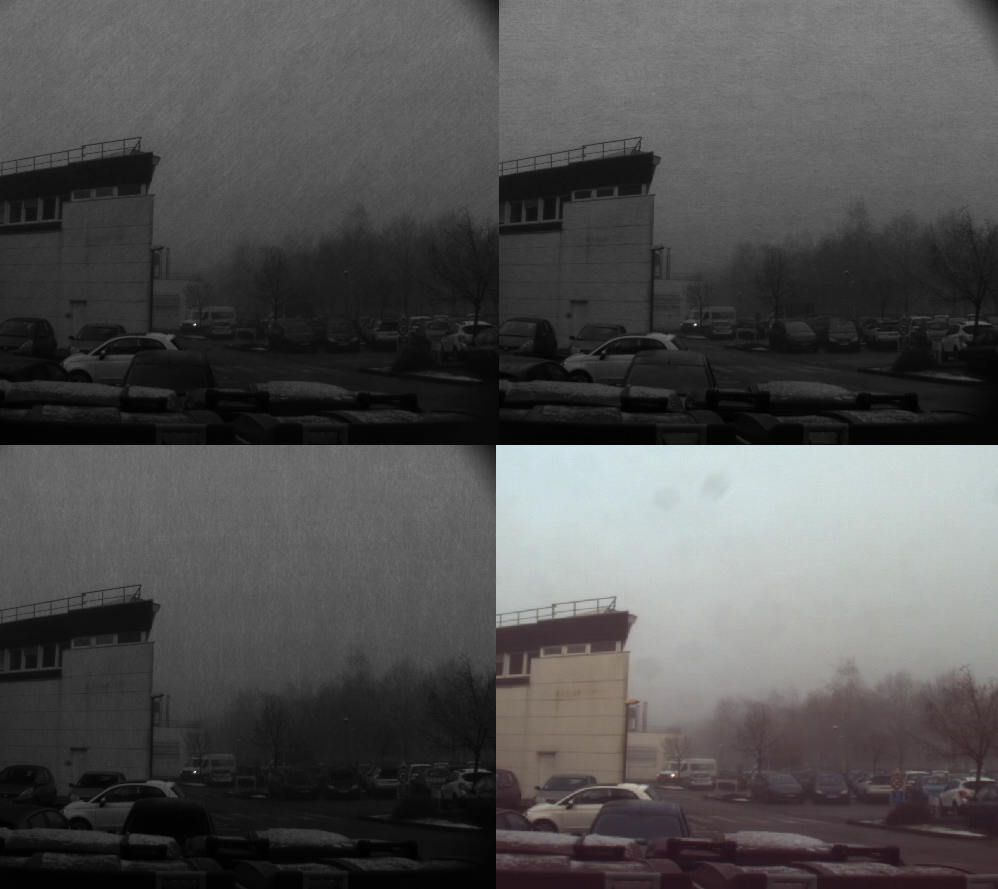}
\caption{On top left $I_0$, on top right $I_{45}$, on bottom left $I_{135}$ and on bottom right the equivalent of this scene in RGB}
\label{fig:3}
\end{figure}

Our final dataset contains 2730 labelled polarization images  divided into  2221  images for the training set and  509 for the testing set. Those images contain about 23K labelled objects including about 90\% objects of the class 'cars' for both the training and the testing sets and 10\% objects of the class 'person'. Table~\ref{tab:1} sums up the number of labelled objects in each class for the training and testing sets. There are less than 30 objects for each of the classes 'bike' and 'motorbike' that is the reason why  theses classes were not considered in the present study. In order to compare the performances with the detection on RGB images, there's also a testing set labelled containing 509 RGB images with the equivalent of the polarization images. Figure~\ref{fig:3} shows an example of those two testing sets. As it was pointed out earlier, the training set only contains images that were taken in sunny days and the testing set only contains images that were taken in foggy days. With this configuration, it was possible to see if polarization images, characterizing an object by its reflection and not only by its shape or intensity, could overcome classical image detection when the weather conditions are not optimal. The MS COCO dataset contains the same configuration as in our dataset, including only few images in adverse weather conditions which are not enough to enable the network to properly detect objects in such conditions. By taking only images in good weather conditions for the training purpose, we were on the same basis. We could then compare the results of classical images detection with RetinaNet-50 and SSD300 trained on the MS COCO dataset and the detection results on polarization images  with the same networks after fine tuning on the polarimetric dataset.

\begin{table}
\begin{center}
\begin{tabular}{c|c|c}
    \specialrule{.2em}{.1em}{.1em} 
    Class name & Training set & Testing set \\
    \specialrule{.2em}{.1em}{.1em} 
    car & 11687 & 9265 \\
    person & 1488 & 442 \\
    bike & 4 & 12 \\
    motorbike & 21 & 0 \\
\end{tabular}
\caption{Number of labelled objects in the database}
\label{tab:1}
\end{center}
\end{table}

\subsection{DATA ENCODING}

Referring to the polarization formalism, the AOP and DOP 
values lie in the following intervals: 
\\
\\
\bigskip\noindent
$\hfill AOP \in \left[\dfrac{-\pi}{2};\dfrac{\pi}{2}\right], \hfill DOP \in [0,1]. \hfill$

To avoid the neural network to get confused by different data formats, each parameter were normalized between 0 and 255. The polarimetric channels could get values in the same format than the RGB  and thus be processed in the same way.

\section{EXPERIMENTS}
After collecting, sorting and labelling all the data needed, it was possible to start training the networks.

\subsection{EXPERIMENTAL SETUP}

\begin{figure}
\centering
\includegraphics[width=0.47\textwidth]{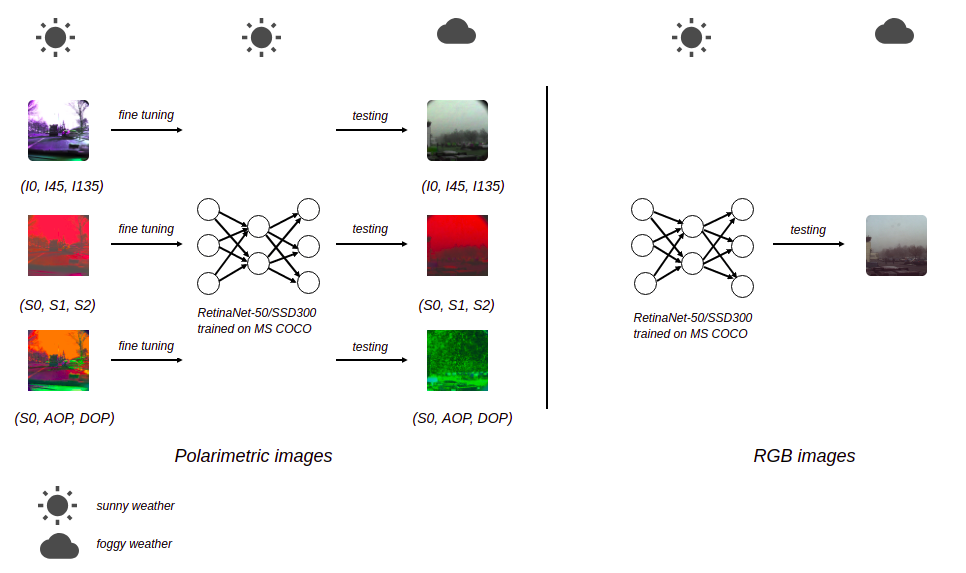}
\caption{Experimental setup}
\label{fig:4}
\end{figure}

To remind the experiments conditions, a model of RetinaNet-50 and a model of SSD300 using a VGG16 as a backbone are used. Both of them are pre-trained on the MS COCO dataset. In order to improve the detection on polarization images and compare the detection with the one in classical images, the cited models were fine tuned on the polarimetric database using the MS COCO dataset as a basis. Because the polarization channels combinations presented in the introduction (i.e. ($I_0$, $I_{45}$, $I_{135}$), ($ S_{0}$, $S_{1}$, $S_{2} $) and ($S_{0}$, AOP, DOP)) contain different information, the networks were fine tuned on each combination separately. The MS COCO dataset is a good basis to fine tune the networks as it contains different road scenes classes including the 4 classes of the polarimetric dataset. Figure~\ref{fig:4} summarizes up the experimental setup.

Because fine tuning requires a very low learning rate to be able to learn the new parameters efficiently, RetinaNet-50 was trained on 50 epoch with a learning rate of $10^{-5}$ and SSD300 on 100 epochs with a learning rate of $10^{-5}$. The Adam optimizer \cite{kingma2014adam} with a $10^{-5}$ learning rate was used for both networks. It is important to note that 50 and 100 epochs for respectively RetinaNet-50 and SSD300 are fixed to make sure the network would converge at the end of the training. The optimal weights are found according to the loss value. 

\subsection{RESULTS AND DISCUSSION}
As a reminder, the networks were trained on a database containing only road scenes when the weather is sunny in order to be synchronized with the MS COCO dataset. Because of the low number of samples in classes 'bike' and 'motorbike' they were skipped in this experiment. The mean average precision $mAP^{d}$ for the data format $d \in$ \{RGB, ($I_0$, $I_{45}$, $I_{135}$), ($S_{0}$, $S_{1}$, $S_{2}$), ($S_0$, AOP, DOP)\} used in this work is given by:
$$
    mAP^{d} = \frac{n_{p} \times AP_{p}^{d}+n_{c} \times AP_{c}^{d}}{n_{p}+n_{c}},
$$
where $AP_{p}^{d}$ and $AP_{c}^{d}$ are the average precision respectively for the classes 'person' and 'car' for the related data format $d$ while $n_{p}$ and $n_{c}$ are the number of instances in the testing set for respectively the classes 'person' and 'car'.

After fine tuning, the $mAP$ was computed for each of the polarization channels combination using the updated weights for each one of them. For RGB, the $mAP$ of the detection from the RetinaNet-50 and the SSD300 trained on the MS COCO dataset was used.

\begin{table}
\begin{center}
\begin{tabular}{c c c c}
    \specialrule{.3em}{.2em}{.2em}
    Entries & Class name & AP no FT & AP FT \\
    \specialrule{.3em}{.2em}{.2em}
    RGB & person & \color{blue}{0.8254} & X \\
        & car & \color{blue}{0.6639} & X \\
    \multicolumn{2}{c}{$mAP$} & \color{blue}{0.6706} & X \\
    \specialrule{.2em}{.1em}{.1em} 
    ($I_0$, $I_{45}$, $I_{135}$) & person & \textbf{0.8556} & \textbf{0.9079} \\
        & car & 0.6064 & \textbf{0.7290} \\
    \multicolumn{2}{c}{$mAP$} & 0.6177 & \color{green}{\textbf{0.7371}} \\
    \specialrule{.2em}{.1em}{.1em} 
    ($S_{0}$, $S_{1}$, $S_{2}$) & person & 0.6945 & \textbf{0.8969} \\
        & car & 0.4114 & \textbf{0.7375} \\
    \multicolumn{2}{c}{$mAP$} & 0.4243 & \color{green}{\textbf{0.7448}} \\
    \specialrule{.2em}{.1em}{.1em} 
    ($S_{0}$, AOP, DOP) & person & 0.0166 & 0.3585 \\
        & car & 0.1265 & 0.6050 \\
    \multicolumn{2}{c}{$mAP$} & 0.1215 & 0.5938\\
\end{tabular}
\end{center}
\caption{Comparison of the detection with RetinaNet-50 before and after fine tuning. AP no FT and AP FT stand respectively for Average Precision before Fine Tuning and Average Precision after Fine Tuning.}
\label{tab:2}
\end{table}

As it can be seen in Table~\ref{tab:2}, when dealing with the polarization channels combinations without fine tuning, the network fails to detect all the objects in the road scene. After fine tuning, the detection with RetinaNet-50 in two polarization channels combinations (($I_0$, $I_{45}$, $I_{135}$) and ($S_{0}$, $S_{1}$, $S_{2}$)) overcome the classical RGB detection when it comes to car and pedestrian detection. However, regarding the detection with SSD300 after fine tuning, it overcomes the classical RGB detection for car as well as for person detection with only one polarization channels combination, ($S_{0}$, $S_{1}$, $S_{2}$).

The percentage of error rate evolution $ER_o^d$ for the object $o \in \{'person', 'car'\}$ and the data format $d$ is given by:

$$
    ER_o^{d} = \frac{1 - AP_o^{RGB} - (1 - AP_o^{d})}{1- AP_o^{RGB}} \times 100,
$$
where $AP_o^{RGB}$ is the average precision for object $o$ with the RGB data format while $AP_o^{d}$ denotes the average precision on the object $o$ and the related data format $d$.

For ($I_0$, $I_{45}$, $I_{135}$), the error rate decreased is of 21.90\% for the car detection and of 47.25\% for the person detection, which are important improvements in term of object detection.\\

SSD300 is known for its bad detection of small objects unlike RetinaNet-50. On top of that, data augmentation is very important to enable SSD300 to learn the new features correctly. This alternative solution was not possible in the polarization context because data augmentation does not preserve the physical interpretation of the scene it represents. Based on this observation, data augmentation was not used in the training phase. As a consequence, SSD300's architecture might not be adapted to properly learn the object polarimetric features. Polarimetric imaging could thus be a real added value, especially for small objects detection and adverse weather conditions. When objects are too small to be detected from their shape or altered due to bad visibility, they can be characterized according to their reflection, which doesn't change with the object's size or occlusion. By learning these new features, the network is still able to detect the objects in altered conditions thanks to the polarimetric property of the reflection.

\begin{figure}
\centering
 \begin{subfigure}{8cm}
    \includegraphics[width=8cm]{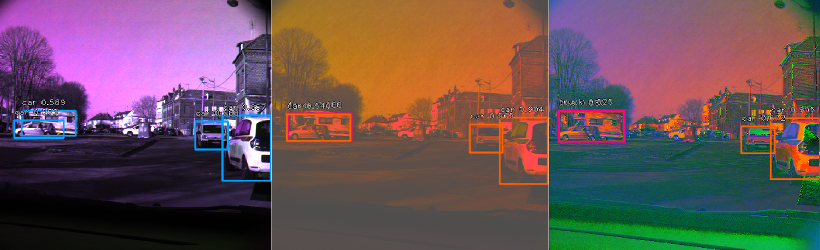}
    \caption{Detection results before fine tuning on ($I_0$, $I_{45}$, $I_{135}$), ($S_{0}$, $S_{1}$, $S_{2}$) and ($ S_{0}$, AOP, DOP)}
\label{fig:6.a}
 \end{subfigure}
 
 \begin{subfigure}{8cm}
    \includegraphics[width=8cm]{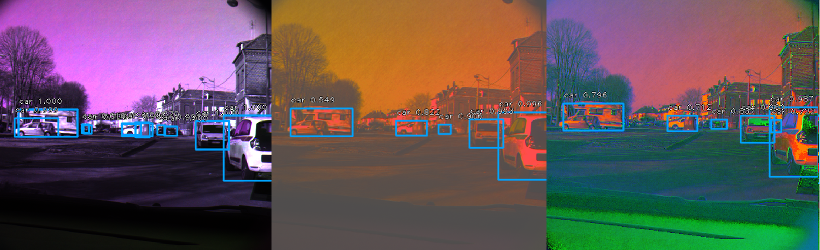}
    \caption{Detection results after fine tuning on ($I_0$, $I_{45}$, $I_{135}$), ($S_{0}$, $S_{1}$, $S_{2}$) and ($ S_{0}$, AOP, DOP)}
\label{fig:6.b}
 \end{subfigure}
 \caption{Detection results with RetinaNet-50}
\end{figure}

Figure~\ref{fig:6.a} shows results of RetinaNet-50 detection before fine tuning on the polarization channels combinations and Figure~\ref{fig:6.b} the detection with the same neural network on the same channels combinations but after fine tuning. The illustrated results prove that fine tuning enabled RetinaNet-50 to detect efficiently objects on polarimetric road scenes images. The network learned successfully these new physical features.

\begin{figure}
\centering
\includegraphics[width=0.47\textwidth]{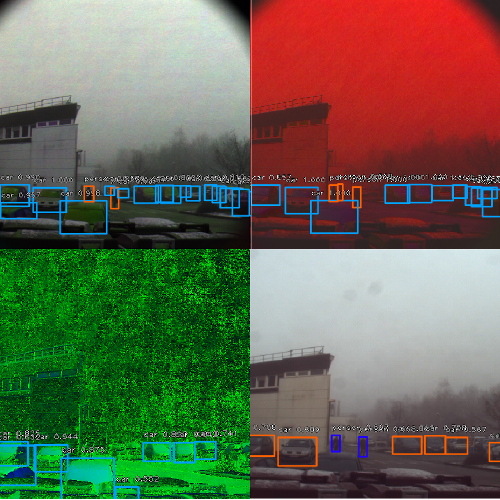}
\caption{Comparison of the detection in foggy weather in the same scene with the different parameters. On top left ($I_0$, $I_{45}$, $I_{135}$), on top right ($ S_{0}$, $S_{1}$, $S_{2} $), on bottom left ($S_{0}$, AOP, DOP) and on bottom right RGB. For the polarization channels combinations detection, blue boxes refer to car and orange boxes refer to pedestrian and for the RGB detection, orange boxes refer to cars and purple boxes refer to pedestrian.}
\label{fig:7}
\end{figure}

The performed experiments and the obtained results show that polarimetric imaging is a real asset in the field of object detection in road scenes. As a matter of fact, this experimental setup showed that even if the network wasn't trained on scenes in adverse weather conditions for both RGB and polarimetric detection purposes, the detection on polarization images achieved better results. An illustration of those performances can be found in Figure~\ref{fig:7}. Polarimetric parameters describe an object features regarding its reflection. This new physical property takes over in the detection process when the shape or the intensity are difficult to detect.

\section{CONCLUSION AND FUTURE WORK}

This paper proves that polarimetric imaging is a real added value in the field of object detection in road scenes. Polarization images associated with deep networks are able to efficiently detect objects in the scene even in case of adverse weather or in presence of small objects.

In the future, the polarimetric database will be increased so as to get more objects in the underrepresented categories (bike and motorbike) and for different weather conditions.
Having a large dataset of polarization images and its counterpart in RGB, would enable to achieve fine tuning on both of them and make a stronger comparison for object detection. It would also be interesting to achieve data augmentation on this dataset in order to reinforce the learning while keeping the polarimetric physical meaning. A fusion scheme of polarization channels with RGB images merits to be studied thoroughly to enhance the average precision of both of them separately.

\addtolength{\textheight}{-12cm}    




\section*{ACKNOWLEDGMENT}

This work is supported by the ICUB project 2017 ANR program : ANR-17-CE22-0011. We also thank our colleagues from the Criann who provided us some computation resources with Myria and by so enabled us to get our results efficiently and faster.

\bibliographystyle{IEEEtran}
\bibliography{bibliography}

\begin{thebibliography}{10}
\providecommand{\url}[1]{#1}
\csname url@rmstyle\endcsname
\providecommand{\newblock}{\relax}
\providecommand{\bibinfo}[2]{#2}
\providecommand\BIBentrySTDinterwordspacing{\spaceskip=0pt\relax}
\providecommand\BIBentryALTinterwordstretchfactor{4}
\providecommand\BIBentryALTinterwordspacing{\spaceskip=\fontdimen2\font plus
\BIBentryALTinterwordstretchfactor\fontdimen3\font minus
  \fontdimen4\font\relax}
\providecommand\BIBforeignlanguage[2]{{%
\expandafter\ifx\csname l@#1\endcsname\relax
\typeout{** WARNING: IEEEtran.bst: No hyphenation pattern has been}%
\typeout{** loaded for the language `#1'. Using the pattern for}%
\typeout{** the default language instead.}%
\else
\language=\csname l@#1\endcsname
\fi
#2}}

\bibitem{yoffie2014mobileye}
D.~B. Yoffie, ``Mobileye: The future of driverless cars,'' \emph{Harvard
  Business School Case}, pp. 715--421, 2014.

\bibitem{hautiere2014enhanced}
N.~Hauti{\`e}re, J.-P. Tarel, H.~Halmaoui, R.~Br{\'e}mond, and D.~Aubert,
  ``Enhanced fog detection and free-space segmentation for car navigation,''
  \emph{Machine vision and applications}, vol.~25, no.~3, pp. 667--679, 2014.

\bibitem{babari2012visibility}
R.~Babari, N.~Hauti{\`e}re, {\'E}.~Dumont, N.~Paparoditis, and J.~Misener,
  ``Visibility monitoring using conventional roadside cameras--emerging
  applications,'' \emph{Transportation research part C: emerging technologies},
  vol.~22, pp. 17--28, 2012.

\bibitem{miron2015evaluation}
A.~Miron, A.~Rogozan, S.~Ainouz, A.~Bensrhair, and A.~Broggi, ``An evaluation
  of the pedestrian classification in a multi-domain multi-modality setup,''
  \emph{Sensors}, vol.~15, no.~6, pp. 13\,851--13\,873, 2015.

\bibitem{bertozzi2006low}
M.~Bertozzi, A.~Broggi, M.~Felisa, G.~Vezzoni, and M.~Del~Rose, ``Low-level
  pedestrian detection by means of visible and far infra-red tetra-vision,'' in
  \emph{2006 IEEE Intelligent Vehicles Symposium}.\hskip 1em plus 0.5em minus
  0.4em\relax IEEE, 2006, pp. 231--236.

\bibitem{morel2006active}
O.~Morel, C.~Stolz, F.~Meriaudeau, and P.~Gorria, ``Active lighting applied to
  three-dimensional reconstruction of specular metallic surfaces by
  polarization imaging,'' \emph{Applied optics}, vol.~45, no.~17, pp.
  4062--4068, 2006.

\bibitem{novikova2018mueller}
T.~Novikova, J.~Rehbinder, J.~Vizet, A.~Pierangelo, R.~Ossikovski, A.~Nazac,
  A.~Benali, and P.~Validire, ``Mueller polarimetry as a tool for optical
  biopsy of tissue,'' in \emph{2018 International Conference Laser Optics
  (ICLO)}.\hskip 1em plus 0.5em minus 0.4em\relax IEEE, 2018, pp. 553--553.

\bibitem{rankin2010passive}
A.~L. Rankin and L.~H. Matthies, ``Passive sensor evaluation for unmanned
  ground vehicle mud detection,'' \emph{Journal of Field Robotics}, vol.~27,
  no.~4, pp. 473--490, 2010.

\bibitem{fan2018polarization}
W.~Fan, S.~Ainouz, F.~Meriaudeau, and A.~Bensrhair, ``Polarization-based car
  detection,'' in \emph{2018 25th IEEE International Conference on Image
  Processing (ICIP)}.\hskip 1em plus 0.5em minus 0.4em\relax IEEE, 2018, pp.
  3069--3073.

\bibitem{kamann2018automotive}
A.~Kamann, P.~Held, F.~Perras, P.~Zaumseil, T.~Brandmeier, and U.~T. Schwarz,
  ``Automotive radar multipath propagation in uncertain environments,'' in
  \emph{2018 21st International Conference on Intelligent Transportation
  Systems (ITSC)}.\hskip 1em plus 0.5em minus 0.4em\relax IEEE, 2018, pp.
  859--864.

\bibitem{wolff1995polarization}
L.~B. Wolff and A.~G. Andreou, ``Polarization camera sensors,'' \emph{IVC},
  vol.~13, no.~6, pp. 497--510, 1995.

\bibitem{blanchon2019outdoor}
M.~Blanchon, O.~Morel, Y.~Zhang, R.~Seulin, N.~Crombez, and D.~Sidib{\'e},
  ``Outdoor scenes pixel-wise semantic segmentation using polarimetry and fully
  convolutional network,'' in \emph{14th International Conference on Computer
  Vision Theory and Applications (VISAPP 2019)}, 2019.

\bibitem{girshick2014rich}
R.~Girshick, J.~Donahue, T.~Darrell, and J.~Malik, ``Rich feature hierarchies
  for accurate object detection and semantic segmentation,'' in
  \emph{Proceedings of the IEEE conference on computer vision and pattern
  recognition}, 2014, pp. 580--587.

\bibitem{girshick2015fast}
R.~Girshick, ``Fast r-cnn,'' in \emph{Proceedings of the IEEE international
  conference on computer vision}, 2015, pp. 1440--1448.

\bibitem{ren2015faster}
S.~Ren, K.~He, R.~Girshick, and J.~Sun, ``Faster r-cnn: Towards real-time
  object detection with region proposal networks,'' in \emph{Advances in neural
  information processing systems}, 2015, pp. 91--99.

\bibitem{everingham2010pascal}
M.~Everingham, L.~Van~Gool, C.~K. Williams, J.~Winn, and A.~Zisserman, ``The
  pascal visual object classes (voc) challenge,'' \emph{International journal
  of computer vision}, vol.~88, no.~2, pp. 303--338, 2010.

\bibitem{Redmon_2016_CVPR}
J.~Redmon, S.~Divvala, R.~Girshick, and A.~Farhadi, ``You only look once:
  Unified, real-time object detection,'' in \emph{The IEEE Conference on
  Computer Vision and Pattern Recognition (CVPR)}, June 2016.

\bibitem{liu2016ssd}
W.~Liu, D.~Anguelov, D.~Erhan, C.~Szegedy, S.~Reed, C.-Y. Fu, and A.~C. Berg,
  ``Ssd: Single shot multibox detector,'' in \emph{European conference on
  computer vision}.\hskip 1em plus 0.5em minus 0.4em\relax Springer, 2016, pp.
  21--37.

\bibitem{Lin2017FocalLF}
T.-Y. Lin, P.~Goyal, R.~B. Girshick, K.~He, and P.~Dollar, ``Focal loss for
  dense object detection,'' \emph{2017 IEEE International Conference on
  Computer Vision (ICCV)}, pp. 2999--3007, 2017.

\bibitem{redmon2017yolo9000}
J.~Redmon and A.~Farhadi, ``Yolo9000: better, faster, stronger,'' in
  \emph{Proceedings of the IEEE conference on computer vision and pattern
  recognition}, 2017, pp. 7263--7271.

\bibitem{redmon2018yolov3}
------, ``Yolov3: An incremental improvement,'' \emph{arXiv preprint
  arXiv:1804.02767}, 2018.

\bibitem{geiger2012we}
A.~Geiger, P.~Lenz, and R.~Urtasun, ``Are we ready for autonomous driving? the
  kitti vision benchmark suite,'' in \emph{2012 IEEE Conference on Computer
  Vision and Pattern Recognition}.\hskip 1em plus 0.5em minus 0.4em\relax IEEE,
  2012, pp. 3354--3361.

\bibitem{janai2017computer}
J.~Janai, F.~G{\"u}ney, A.~Behl, and A.~Geiger, ``Computer vision for
  autonomous vehicles: Problems, datasets and state-of-the-art,'' \emph{arXiv
  preprint arXiv:1704.05519}, 2017.

\bibitem{bass1995handbook}
M.~Bass, E.~W. Van~Stryland, D.~R. Williams, and W.~L. Wolfe, \emph{Handbook of
  optics}.\hskip 1em plus 0.5em minus 0.4em\relax McGraw-Hill New York, 1995,
  vol.~2.

\bibitem{born2013principles}
M.~Born and E.~Wolf, \emph{Principles of optics: electromagnetic theory of
  propagation, interference and diffraction of light}.\hskip 1em plus 0.5em
  minus 0.4em\relax Elsevier, 2013.

\bibitem{ainouz2013adaptive}
S.~Ainouz, O.~Morel, D.~Fofi, S.~Mosaddegh, and A.~Bensrhair, ``Adaptive
  processing of catadioptric images using polarization imaging: towards a
  pola-catadioptric model,'' \emph{Optical engineering}, vol.~52, no.~3, p.
  037001, 2013.

\bibitem{lin2014}
T.-Y. Lin, M.~Maire, S.~Belongie, J.~Hays, P.~Perona, D.~Ramanan, P.~Dollar,
  and C.~L. Zitnick, ``Microsoft coco: Common objects in context,'' in
  \emph{ECCV}.\hskip 1em plus 0.5em minus 0.4em\relax Springer, 2014, pp.
  740--755.

\bibitem{He_2016_CVPR}
K.~He, X.~Zhang, S.~Ren, and J.~Sun, ``Deep residual learning for image
  recognition,'' in \emph{The IEEE Conference on Computer Vision and Pattern
  Recognition (CVPR)}, June 2016.

\bibitem{simonyan2014very}
K.~Simonyan and A.~Zisserman, ``Very deep convolutional networks for
  large-scale image recognition,'' \emph{arXiv preprint arXiv:1409.1556}, 2014.

\bibitem{yu2018bdd100k}
F.~Yu, W.~Xian, Y.~Chen, F.~Liu, M.~Liao, V.~Madhavan, and T.~Darrell,
  ``Bdd100k: A diverse driving video database with scalable annotation
  tooling,'' \emph{arXiv preprint arXiv:1805.04687}, 2018.

\bibitem{kingma2014adam}
D.~P. Kingma and J.~Ba, ``Adam: A method for stochastic optimization,''
  \emph{arXiv preprint arXiv:1412.6980}, 2014.

\end{thebibliography}

\end{document}